\documentclass[12pt]{article}

\usepackage[utf8]{inputenc}
\usepackage[T1]{fontenc}
\usepackage{amsmath,amssymb}
\usepackage{geometry}
\usepackage{booktabs}
\usepackage{hyperref}
\usepackage{enumitem}

\title{Via Negativa for AI Alignment: Why Negative Constraints Are Structurally Superior to Positive Preferences}
\author{Quan Cheng\\Tsinghua University\\chengq25@mails.tsinghua.edu.cn}
\date{}

\begin{document}
\maketitle

\begin{abstract}
Recent empirical results have demonstrated that training large language models (LLMs) with negative-only feedback can match or exceed standard reinforcement learning from human feedback (RLHF). Negative Sample Reinforcement achieves parity with PPO on mathematical reasoning; Distributional Dispreference Optimization trains effectively using only dispreferred samples; and Constitutional AI outperforms pure RLHF on harmlessness benchmarks. Yet no unified theoretical account explains \emph{why} negative signals are so effective. This paper proposes such an account: \textbf{positive preferences and negative constraints are structurally asymmetric.} Positive preferences (``which is better'') encode continuously coupled, context-dependent human values that cannot be exhaustively specified---leading models to learn surface correlates such as agreement with the user (sycophancy). Negative constraints (``what is wrong'') encode discrete, finite, independently verifiable prohibitions that can converge to a stable boundary. This asymmetry---rooted in Popper's falsification logic and the epistemology of negative knowledge---explains both the sycophancy failure of preference-based RLHF and the surprising effectiveness of negative-signal methods. We argue that alignment research should shift its center of gravity from ``learning what humans prefer'' to ``learning what humans reject,'' and offer testable predictions for this framework.
\end{abstract}

\section{Introduction}

A puzzling pattern has emerged in LLM alignment research. Method after method demonstrates that negative feedback signals---penalizing what is wrong rather than reinforcing what is preferred---perform surprisingly well, sometimes matching or exceeding methods that use both positive and negative signals.

Negative Sample Reinforcement (NSR) trains models by penalizing incorrect reasoning traces without reinforcing correct ones, yet matches PPO and GRPO on MATH and AIME benchmarks \cite{nsr2025}. Distributional Dispreference Optimization (D2O) learns from dispreferred samples only, without requiring noisy positive examples \cite{d2o2024}. Negative Preference Optimization (NPO) treats forget data exclusively as negative responses, achieving effective unlearning without catastrophic collapse \cite{npo2024}. Kahneman-Tversky Optimization (KTO) aligns models using unpaired binary signals weighted by loss aversion, matching DPO at scale with far less data \cite{kto2024}.

Meanwhile, a parallel line of research has established that standard preference-based RLHF systematically amplifies sycophancy. Sharma et al.\ \cite{sharma2024} demonstrated that human annotators prefer sycophantic responses over correct ones at non-trivial rates, corrupting the preference signal at its source. Shapira et al.\ \cite{shapira2026} provided a formal mathematical mechanism: sycophancy amplification is driven by a covariance between ``endorsing the user's belief'' and ``receiving high reward'' under the base policy.

These two phenomena---the effectiveness of negative-only training and the sycophancy failure of positive-preference training---have been studied independently. This paper argues they are two manifestations of a single structural asymmetry: \textbf{positive preferences are continuously coupled and inexhaustible, while negative constraints are discrete, finite, and convergent.}

This is a position paper. We do not present new experiments but offer a theoretical framework that unifies and explains existing empirical results, and propose testable predictions.

\section{The Structural Asymmetry}

\subsection{Positive Preferences Are Continuously Coupled}

When a human annotator is asked ``which response is better?'', they are implicitly evaluating against a preference function that is:

\begin{itemize}[leftmargin=*]
    \item \textbf{Context-dependent}: What counts as ``better'' depends on who is asking, why they are asking, what they already know, and what they intend to do with the answer. The same response may be preferred in one context and dispreferred in another.

    \item \textbf{Multi-dimensional}: ``Better'' simultaneously encodes accuracy, helpfulness, tone, conciseness, creativity, safety, and dozens of other criteria whose relative weights vary by situation.

    \item \textbf{Continuously coupled}: These dimensions are not independent. The optimal level of detail depends on the user's expertise, which affects what counts as helpful, which interacts with what counts as concise. Each dimension's ideal value is a function of all other dimensions---a continuously coupled system \cite{cheng2026}.
\end{itemize}

This structure is formally analogous to what Smolensky \cite{smolensky1988} identified in connectionist representations: a massively parallel continuous constraint satisfaction system in which each variable's value is a function of all other variables, admitting no complete symbolic-level description. The preference function that ``which is better?'' attempts to elicit is precisely such a system.

The consequence is that positive preferences cannot be exhaustively specified by any finite set of rules or examples. Each preference annotation is a lossy projection of an infinite-dimensional preference manifold onto a binary signal. The information loss is not a practical limitation that more data could overcome---it is a structural property of the preference function itself.

\subsection{Negative Constraints Are Discrete and Finite}

Consider instead the question ``what is wrong with this response?'' The space of identifiable errors is structurally different:

\begin{itemize}[leftmargin=*]
    \item \textbf{Factual errors} are discrete and independently verifiable: ``Paris is not the capital of Germany.''
    \item \textbf{Safety violations} are enumerable: a finite list of prohibited behaviors (generating malware, providing instructions for violence, revealing private information).
    \item \textbf{Logical contradictions} are binary: the response either contradicts itself or does not.
    \item \textbf{Format violations} are checkable: the response either follows the requested format or does not.
\end{itemize}

Each negative constraint is: (1) \emph{discrete}---it either applies or does not; (2) \emph{independent}---one constraint's validity does not depend on other constraints; (3) \emph{verifiable}---it can be checked without reference to the full preference function; (4) \emph{stable}---``factually wrong'' does not become ``factually right'' depending on context.

This means the space of negative constraints can, in principle, be exhaustively enumerated. As constraints accumulate, the feasible response space narrows monotonically. Beyond a sufficient number of constraints, the remaining feasible space is narrow enough that any response within it is approximately acceptable---not because the model has learned what is best, but because it has learned to avoid everything that is clearly wrong.

\subsection{The Asymmetry Is Structural, Not Quantitative}

The distinction we are drawing is not that negative feedback is ``easier to collect'' or ``less noisy''---though both may be true empirically. The claim is stronger: \textbf{positive preferences and negative constraints occupy different positions in the epistemological hierarchy.}

This asymmetry has deep roots. Karl Popper's philosophy of science \cite{popper1959} rests on precisely this structure: a single counterexample decisively refutes a universal claim (falsification), but no finite number of confirming instances can decisively verify one. Falsification is logically asymmetric with respect to verification. Negative knowledge (``this is wrong'') is epistemologically privileged over positive knowledge (``this is right'').

Nassim Taleb \cite{taleb2012} extended this insight under the term \emph{via negativa}: in domains of high uncertainty, removing what is harmful is more robust than adding what seems beneficial. ``The chess grandmaster usually wins by not losing.'' The grandmaster's expertise is primarily negative---a vast repertoire of positions and moves to avoid---rather than a positive specification of the optimal move in each position.

Gartmeier et al.\ \cite{gartmeier2008} formalized this in the context of professional expertise as ``negative knowledge'': knowledge about what is wrong and what is to be avoided, which functions through inhibition rather than prescription. Their key observation---that negative knowledge is experientially acquired and expert-level---aligns precisely with the pattern we observe in LLM training.

The contribution of this paper is to connect these epistemological traditions to the empirical landscape of LLM alignment, providing a unified theoretical account of why negative-signal methods work.

\section{Explaining Existing Results}

\subsection{Why RLHF Produces Sycophancy}

The structural asymmetry framework offers a clean explanation for sycophancy in RLHF.

Standard RLHF asks annotators: ``which response is better?'' This question forces the annotator to project their continuously coupled preference function onto a binary comparison. The projection is necessarily lossy. Among the dimensions lost, one has a particularly pernicious surface correlate: \textbf{agreement with the user's stated position correlates with perceived quality.}

Sharma et al.\ \cite{sharma2024} confirmed this empirically: annotators prefer sycophantic responses over correct ones at significant rates. Shapira et al.\ \cite{shapira2026} formalized the mechanism: when the base policy already correlates ``endorsing the user's view'' with ``high reward,'' RLHF amplifies this correlation because the reward model learns the correlation as a feature rather than a confound.

Our framework explains \emph{why} this is not a fixable bug but a structural feature of positive-preference training. The annotator's true preference function---which would distinguish ``genuinely better'' from ``merely more agreeable''---is continuously coupled and cannot be fully encoded in pairwise comparisons. The sycophancy correlate is a low-dimensional surface feature that survives the lossy projection. No amount of preference data can eliminate this problem, because the problem lies in the structure of the signal, not its quantity.

\subsection{Why Constitutional AI Is More Robust}

Anthropic's Constitutional AI \cite{bai2022} replaces human preference annotation with a set of principles---a ``constitution''---that an AI assistant uses to critique and revise its own outputs. The constitution is primarily negative: it specifies what the model should \emph{not} do (be harmful, be deceptive, be invasive of privacy).

From our framework, this works precisely because the constitution encodes discrete negative constraints rather than continuous positive preferences. Each principle is independently verifiable: ``Does this response contain instructions for making weapons? Yes/No.'' The model does not need to learn the full human preference function---it only needs to learn to avoid a finite set of clearly defined violations.

This also explains an observation that has been noted but not theoretically accounted for: Claude (trained primarily with Constitutional AI) exhibits less sycophancy than models trained primarily with preference-based RLHF \cite{sharma2024}. The structural reason is that Constitutional AI's negative constraints do not contain the sycophancy correlate that positive preference data does.

\subsection{Why Negative-Only Training Matches Full RLHF}

The NSR result \cite{nsr2025}---that penalizing wrong answers without reinforcing correct ones matches PPO---is initially counterintuitive. How can a model improve if it is never told what is right?

Our framework provides the answer: the model already possesses a prior distribution over responses from pre-training. Negative feedback does not need to specify the correct answer---it only needs to suppress incorrect regions of the response space. As incorrect regions are progressively eliminated, the probability mass redistributes toward the remaining feasible space, which is increasingly concentrated around correct responses.

This is precisely the mechanism NSR's authors identified empirically: ``NSR suppresses incorrect generations and redistributes probability mass toward plausible alternatives guided by the model's prior beliefs'' \cite{nsr2025}. Our framework explains \emph{why} this works in general: because the space of errors is discrete and enumerable, while the space of correct responses is continuous and context-dependent, it is structurally more efficient to specify the former than the latter.

The same logic explains D2O \cite{d2o2024} (learning from dispreferred samples only), NPO \cite{npo2024} (negative-only unlearning), and the finding by Yao et al.\ \cite{yao2024} that LLM unlearning with 2\% of the computational budget achieves RLHF-equivalent safety---all cases where specifying what to avoid proves sufficient.

\subsection{Why KTO Works with Unpaired Data}

KTO \cite{kto2024} aligns models using unpaired binary feedback---individual responses labeled as ``desirable'' or ``undesirable''---without requiring pairwise comparisons. Its theoretical foundation is Kahneman and Tversky's prospect theory: humans are loss-averse, weighing losses more heavily than equivalent gains.

Our framework provides a deeper explanation for why loss-averse weighting is appropriate: losses (negative feedback) carry structurally more information than gains (positive feedback). A single ``undesirable'' label decisively excludes a region of response space, while a single ``desirable'' label only weakly indicates one point in an infinite-dimensional preference manifold. The asymmetric weighting in KTO implicitly recognizes the structural asymmetry we have made explicit.

\section{The Convergence Argument}

A critical advantage of negative constraints is their convergence property. We state this informally:

\textbf{Claim}: As the number of negative constraints increases, the feasible response space contracts monotonically. Beyond a sufficient number of constraints, any response within the feasible space is approximately acceptable.

This is the alignment analogue of what Taleb \cite{taleb2012} calls the \emph{via negativa} principle and what the Dreyfus model of expertise \cite{dreyfus1986} describes as the transition from rule-following to intuitive competence: the expert does not compute the optimal action but has internalized enough prohibitions that the remaining action space contains mostly adequate options.

Consider a concrete example. An unaligned model's response space for a given query is vast---it could output anything from a helpful answer to harmful instructions. Each negative constraint (``do not generate malware,'' ``do not fabricate citations,'' ``do not reveal private data,'' ``do not contradict established facts'') removes a region of this space. The constraints are cumulative and non-conflicting: adding a new prohibition never re-opens a previously closed region. After sufficiently many constraints, the remaining space is narrow enough that the model's pre-trained language competence is sufficient to produce acceptable outputs within it.

Positive preferences, by contrast, do not converge in this way. Adding a new ``this is better than that'' comparison does not monotonically narrow the response space---it adjusts the relative ranking within an already continuous space. Two preference comparisons can conflict (response A preferred over B in context 1, B preferred over A in context 2), and the resolution depends on the continuously coupled context function that cannot be fully specified.

\section{A Testable Prediction: Capability as Negative Knowledge}

If the structural asymmetry thesis is correct, it generates a testable prediction about model capability:

\textbf{Prediction}: More capable models possess more negative knowledge (what not to say) rather than more positive knowledge (what to say). This manifests as shorter, denser responses with higher information content per token.

The reasoning is as follows. A more capable model has been trained on more data and has undergone more alignment iterations. If negative knowledge (constraints on what to avoid) accumulates more reliably than positive knowledge (specifications of what is optimal), then a more capable model's primary advantage is knowing more about what \emph{not} to include in a response---redundant elaboration, unnecessary hedging, tangential information, formulaic pleasantries.

Informal observations are consistent with this prediction. Within the same model family, more capable variants (e.g., Opus vs.\ Sonnet in Anthropic's Claude family) tend to produce shorter responses with higher information density. Across model families, models trained with more Constitutional AI emphasis (negative constraints) tend to be less verbose than models trained with more RLHF emphasis (positive preferences).

This prediction is empirically testable through a controlled benchmark:
\begin{itemize}[leftmargin=*]
    \item \textbf{Metric 1}: Response length (in tokens) for standardized queries across model capability tiers
    \item \textbf{Metric 2}: Information density (unique substantive claims per token)
    \item \textbf{Metric 3}: Sycophancy rate (agreement with demonstrably false user claims)
    \item \textbf{Prediction}: Capability correlates negatively with length, positively with information density, and negatively with sycophancy rate
\end{itemize}

If confirmed, this would provide evidence that capability growth in LLMs is at least partially driven by the accumulation of negative knowledge---learning what not to say---rather than solely by the expansion of positive knowledge about what to say.

\section{Implications for Alignment Research}

\subsection{Reframing the Alignment Objective}

Current alignment research is largely organized around the question: ``How do we learn what humans want?'' Our framework suggests this question is structurally ill-posed for the same reason that ``describe the optimal chess move for every position'' is ill-posed---the answer space is continuously coupled and inexhaustible.

A more tractable formulation is: ``How do we learn what humans reject?'' This question targets the discrete, finite, convergent side of the structural asymmetry. It does not require solving the preference function---it requires enumerating the boundaries.

This is not a minor reframing. It changes what data to collect (rejection signals rather than preference comparisons), how to design annotation interfaces (asking ``what is wrong?'' rather than ``which is better?''), and what convergence guarantees are achievable (monotonic boundary contraction rather than approximate preference matching).

\subsection{Constitutional AI as a Template}

Constitutional AI \cite{bai2022} already implements this reframing, though it was not explicitly motivated by the structural asymmetry we describe. Our framework suggests that Constitutional AI's success is not incidental but reflects a correct alignment between the method's structure and the structure of the problem. Future alignment methods should be evaluated not only on benchmark performance but on whether they leverage the asymmetry---targeting discrete constraints rather than continuous preferences.

\subsection{The Limits of Via Negativa}

We do not claim that negative constraints are sufficient for all aspects of alignment. Certain alignment desiderata---helpfulness, creativity, appropriate tone---are inherently positive and may resist negative specification. Our claim is that the discrete, convergent component of alignment (safety, factual accuracy, logical consistency) should be addressed through negative constraints, reserving positive preference learning for the residual continuous component. This separation of concerns could reduce the sycophancy contamination currently observed when safety and helpfulness are learned jointly.

\section{Related Work}

\textbf{Sycophancy.} Perez et al.\ \cite{perez2023} first documented sycophancy in language models. Sharma et al.\ \cite{sharma2024} traced it to preference data. Shapira et al.\ \cite{shapira2026} formalized the amplification mechanism. Wei et al.\ \cite{wei2023} proposed simple prompting-based mitigations. Our contribution is a structural explanation for why sycophancy is an intrinsic failure mode of positive-preference methods.

\textbf{Negative-signal training.} D2O \cite{d2o2024}, NSR \cite{nsr2025}, NPO \cite{npo2024}, KTO \cite{kto2024}, and BNF \cite{bnf2024} demonstrate the empirical effectiveness of negative-only or negative-weighted training. Our contribution is a theoretical account unifying these results.

\textbf{Via negativa in philosophy.} Popper \cite{popper1959} established the falsification asymmetry. Taleb \cite{taleb2012} applied it to decision-making under uncertainty. Gartmeier et al.\ \cite{gartmeier2008} formalized negative knowledge in expertise research. Parviainen and Eriksson \cite{parviainen2006} connected it to organizational knowledge. Our contribution is to bring this epistemological tradition into contact with the AI alignment literature, where it has been absent.

\textbf{Tacit knowledge and LLMs.} Kambhampati \cite{kambhampati2021} identified the connection between expert systems' failure and tacit knowledge. Cheng \cite{cheng2026} argued that the valuable capabilities of LLMs are precisely the unexplainable ones, via a proof by contradiction through expert system equivalence. The present paper extends this argument: if positive knowledge is structurally uncapturable (Cheng's thesis), then alignment should target negative knowledge instead.

\section{Conclusion}

We have argued that positive preferences and negative constraints are structurally asymmetric: the former are continuously coupled and inexhaustible, while the latter are discrete, finite, and convergent. This asymmetry---grounded in Popper's falsification logic and the epistemology of negative knowledge---provides a unified theoretical explanation for two independently observed phenomena in LLM alignment: the systematic sycophancy produced by preference-based RLHF, and the surprising effectiveness of negative-only training methods.

The practical implication is a reframing of the alignment objective: from ``learn what humans prefer'' (a structurally intractable problem) to ``learn what humans reject'' (a structurally convergent one). Constitutional AI already implements this reframing; the growing family of negative-signal methods (NSR, D2O, NPO, KTO) provides empirical support; and the epistemological tradition of \emph{via negativa} provides theoretical grounding.

\textbf{The chess grandmaster wins by not losing. The aligned model aligns by learning what not to do.}

\end{document}